Keyword for classification: Simulation, Agent-Based Simulation, Multi-Agent Systems, Emergent Behaviour, What-if Analysis, Decision Support.

File name: 051Siebers

# INTRODUCTION TO MULTI-AGENT SIMULATION

Peer-Olaf Siebers * and Uwe Aickelin

School of Computer Science & IT (ASAP)

University of Nottingham

Nottingham, NG8 1BB

UK

Email: pos@cs.nott.ac.uk; uxa@cs.nott.ac.uk

# INTRODUCTION TO MULTI-AGENT SIMULATION

## INTRODUCTION

When designing systems that are complex, dynamic and stochastic in nature, simulation is generally recognised as one of the best design support technologies, and a valuable aid in the strategic and tactical decision making process. A simulation model consists of a set of rules that define how a system changes over time, given its current state. Unlike analytical models, a simulation model is not solved but is run and the changes of system states can be observed at any point in time. This provides an insight into system dynamics rather than just predicting the output of a system based on specific inputs. Simulation is not a decision making tool but a decision support tool, allowing better informed decisions to be made. Due to the complexity of the real world, a simulation model can only be an approximation of the target system. The essence of the art of simulation modelling is abstraction and simplification. Only those characteristics that are important for the study and analysis of the target system should be included in the simulation model.

The purpose of simulation is either to better understand the operation of a target system, or to make predictions about a target system's performance. It can be viewed as an artificial white-room which allows one to gain insight but also to test new theories and practices without disrupting the daily routine of the focal organisation. What you can expect to gain from a simulation study is very well summarised by FIRMA (2000). His idea is that if the theory that has been framed about the target system holds, and if this theory has been adequately translated into a computer model this would allow you to answer some of the following questions:

- Which kind of behaviour can be expected under arbitrarily given parameter combinations and initial conditions?
- Which kind of behaviour will a given target system display in the future?
- Which state will the target system reach in the future?

The required accuracy of the simulation model very much depends on the type of question one is trying to answer. In order to be able to respond to the first question the simulation model needs to be an explanatory model. This requires less data accuracy. In comparison, the simulation model required to answer the latter two questions has to be predictive in nature and therefore needs highly accurate input data to achieve credible outputs. These predictions involve showing trends, rather than giving precise and absolute predictions of the target system performance.

The numerical results of a simulation experiment on their own are most often not very useful and need to be rigorously analysed with statistical methods. These results then need to be considered in the context of the real system and interpreted in a qualitative way to make meaningful recommendations or compile best practice guidelines. One needs a good working knowledge about the behaviour of the real system to be able to fully exploit the understanding gained from simulation experiments.

The goal of this chapter is to brace the newcomer to the topic of what we think is a valuable asset to the toolset of analysts and decision makers. We will give you a summary of information we have gathered from the literature and of the experiences that we have made first hand during the last five years, whilst obtaining a better understanding of this exciting technology. We hope that this will help you to avoid some pitfalls that we have unwittingly encountered. Section 2 is an introduction to the different types of simulation used in Operational Research and Management Science with a clear focus on agent-based simulation.

In Section 3 we outline the theoretical background of multi-agent systems and their elements to prepare you for Section 4 where we discuss how to develop a multi-agent simulation model. Section 5 outlines a simple example of a multi-agent system. Section 6 provides a collection of resources for further studies and finally in Section 7 we will conclude the chapter with a short summary.

SIMULATION TECHNIQUES

Operational Research usually employs three different types of simulation modelling to help understand the behaviour of organisational systems, each of which has its distinct application area: Discrete Event Simulation (DES), System Dynamics (SD) and Agent Based Simulation (ABS). DES models a system as a set of entities being processed and evolving over time according to the availability of resources and the triggering of events. The simulator maintains an ordered queue of events. DES is widely used for decision support in manufacturing (batch and process) and service industries. SD takes a top down approach by modelling system changes over time. The analyst has to identify the key state variables that define the behaviour of the system and these are then related to each other through coupled, differential equations. SD is applied where individuals within the system don't have to be highly differentiated and knowledge on the aggregate level is available, for example modelling population, ecological and economic systems. In an ABS model the researcher explicitly describes the decision processes of simulated actors at the micro-level. Structures emerge at the macro level as a result of the actions of the agents, and their interactions with other agents and the environment. Whereas the first two simulation methods are well matured and established in academia as well as in industry, the latter is mainly used as a research tool in academia, for example in the Social Sciences, Economics, Ecology (where it is often referred to as individual-based modelling), and Political Science. Some example applications in these fields can be found in Table 1.

Table 1 Examples of ABS applications

| Field | Application Examples |
|---|---|
| Social Science | Insect societies, group dynamics in fights, growth and decline of ancient societies, group learning, spread of epidemics, civil disobedience |
| Economics | Stock market, self organising markets, trade networks, consumer behaviour, deregulated electric power markets |
| Ecology | Population dynamics of salmon and trout, land use dynamics, flocking behaviour in fish and birds, rain forest growth |
| Political Sciences | Water rights in developing countries, party competition, origins and patterns of political violence, power sharing in multicultural states |

Although computer simulation has been used widely since the 1960s, ABS only became popular in the early 1990s (Epstein & Axtell, 1996). It is now a well established simulation modelling tool in academia and on the way to achieving the same recognition in industry. The history of agent-based modelling is not well documented. This is most likely due to the fact that there is no general consensus about a definition of what deserves to be called an agent and hence opinions in the agent community about the beginnings of agent-based modelling differ. The technical methodology of computational models of multiple interacting agents was initially developed during the 1940s when John von Neumann started to work on cellular automata. A cellular automaton is a set of cells, where each cell can be in one of many pre-defined states, such as forest or farmland. Changes in the state of a cell occur based on the prior states of that cell and the history of its neighbouring cells. Other famous examples of theoretical and abstract early agent developments that show how simple rules can explain macro-level phenomena are Thomas Schelling's study of housing segregation pattern development and Robert Axelrod's prisoner's dilemma tournaments (Janssen & Ostrom, 2006).

Probably the earliest form of agent-type work that has been implemented dates back to the early 1960s when William McPhee published work on modelling voter behaviour (SIMSOC, 2004). In the 1970s academics like Jim Doran and Scott Moss were using agent-based modelling in Social Sciences to address social phenomena such as trade networks, dynamics of settlement patterns, population dynamics, and dynamics of political systems. In the 1980s it was suggested deploying distributed artificial intelligence and multi-actor systems into mainstream Social Science. Other sources suggest that the beginnings of agent-based modelling of real-world social systems can be attributed to Craig Reynolds, modelling the reality of lively biological agents (e.g. bird flocking behaviour) in the mid 1980s, nowadays known as artificial life.

ABS is well suited to modelling systems with heterogeneous, autonomous and pro-active actors, such as human-centred systems. As the relative importance of different industries to the economies of developed countries shift more and more from manufacturing towards the service sector, system analysts are looking for new tools that allow them to take the key element of such systems, such as people and their behaviour, into account when trying to understand and predict the behaviour and performance of these systems. The behaviour of humans differs notably between people, tasks and systems and therefore the heterogeneous and diverse nature of the actors needs to be taken into account during the design and re-design process.

So, what is it that makes ABS particularly well suited to modelling organisations? ABS is a bottom-up approach and is used in situations for which individual variability between the agents cannot be neglected. It allows understanding of how the dynamics of many real systems arise from traits of individuals and their environment. It allows modelling of a heterogeneous population where each agent might have personal motivations and incentives, and to represent groups and group interactions. Multi-agent systems are promising as models of organisations because they are based on the idea that most work in human organisations is done based on intelligence, communication, co-operation, negotiation, and massive parallel processing (Gazendam, 1993). In DES models it is common practice to model people as deterministic resources ignoring their performance variation and their pro-active behaviour. With these simplifications it is not possible to make accurate predictions about the system performance (Siebers, 2006). ABS models allow one to take both into account. Each agent's behaviour is defined by its own set of attribute values which allows to model variation in each individual's behaviour and the simulation design is decentralised which allows the agents to be pro-active. ABS is suited to systems driven by interactions among their entities and can reveal what appears to be complex emergent behaviour at the system level even when the agents involved have fairly simple behaviour.

For the same reasons ABS is extensively used by the game and film industry to develop realistic simulations of individual characters and societies. ABS is used in computer games, for example The SIMS$^{TM}$ (ZDNet, 2000) or in films when diverse heterogeneous characters animations are required, for example the Orcs in Lord of the Rings (BBC, 2005).

One should also be aware that there are also some disadvantages in using ABS. ABS has a higher level of complexity compared to other simulation techniques as all the interactions between agents and between the agent and the environment have to be defined. Furthermore, ABS has high computational requirements.

MULTI-AGENT SYSTEMS

There is a wide range of existing application domains that are making use of the agent paradigm and develop agent-based systems, for example in software technology, robotics, and complex systems. Luck et al. (2005) make a distinction between two main Multi-Agent System (MAS) paradigms: multi-agent decision systems and multi-agent simulation systems. In multi-agent decision systems, agents participating in the system must make joint decisions as a group. Mechanisms for joint decision-making can be based on economic mechanisms, such as an auction, or alternative mechanisms, such as argumentation. In multi-agent simulation systems the MAS is used as a model to simulate some real-world domain. Typical use is in domains involving many different components, interacting in diverse and complex ways and where the system-level properties are not readily inferred from the properties of the components. In this chapter we focus on the latter paradigm and here in particular on the modelling of organisations.

Organisations as Complex Systems

==Complex Systems Science== studies how dynamics of real systems arise from traits of individuals and their environment. It cuts across all traditional disciplines of science, as well as engineering, management, and medicine and is about understanding the indirect effects. Problems that are difficult to solve are often hard to understand because the causes and effects are not obviously related (Bar-Yam, 1997).

==Complex Adaptive Systems== (CAS) are systems that change their behaviour in response to their environment. The adaptive changes that occur are often relevant to achieving a goal or objective. CAS are denoted by the following three characteristics: evolution, aggregate behaviour and anticipation (Holland, 1992). Here, evolution refers to the adaptation of systems to changing environments, aggregate behaviour refers to the emergence of overall system behaviour from the behaviour of its components, and anticipation refers to the expectations the intelligent agents involved have regarding future outcomes. Since CAS adapt to their environment, the effect of environmental change cannot be understood by considering its direct impact alone. Therefore, the indirect effects also have to be considered due to the adaptive response.

Organisations are basically groups of people working together to attain common goals. They can also be characterised as CAS composed of intelligent, task-oriented, boundedly-rational, and socially-situated agents that are faced with an environment that also has the potential for change (Carley & Prietula, 1994). A way of studying such organisations is by use of the toolset of Computational Organisation Theory (COT) which often employs multi-agent based simulation models (Skvoretz, 2003) where the organisation is composed of a number of intelligent agents. The application of these models helps to determine what organisational designs make sense in which situation, and what are the relative costs and benefits of these various configurations that exhibit degrees of equifinality (Carley & Gasser, 1999). With the help of these models new concepts, theories, and knowledge about organising and organisations can be uncovered or tested and then the computational abstractions can be reflected back to actual organisational practice. Unlike traditional multi-agent models, COT models draw on and have embedded in them empirical knowledge from Organization Science about how the human organizations operate and about basic principles for organizing (Carley & Gasser, 1999).

Agents, Intelligence and Multi-Agent Systems

Different disciplines have their own guidelines of what deserves to be called an agent and even in Computer Science there is no consensus about how to define the term agent. However, it is generally accepted that an agent is situated in an environment and has to show autonomous behaviour in order to meet the design objectives (Wooldridge, 2002). Here autonomy means that the agent has control over its own actions and is able to act without human intervention. In order to be regarded as an intelligent agent, the agent has to be capable of flexible autonomous actions. This means the agent needs to be able to master:

- responsive behaviour (perceiving its environment and responds to changes in a timely fashion)
- pro-active behaviour (showing goal-directed behaviour by taking the initiative and is opportunistic)
- social behaviour (interacting with other agents and the environment when appropriate)
- flexible behaviour (having a range of ways to achieve a given goal and being able to recover from failure)

In simple terms an intelligent agent can be described as a discrete autonomous entity with its own goals and behaviours and the capability to interact, and adapt and modify its behaviours. It is important that there is a balance between responsive and goal-directed behaviour. Agents are simply a way of structuring and developing software that offers certain benefits and is very well suited to certain types of applications. In real world applications often agents do not possess all of the capabilities described above, and therefore there is always an argument if these incomplete intelligent agents are entitled to be called intelligent.

A MAS is simply a collection of heterogeneous and diverse intelligent agents that interact with each other and their environment. The interaction can be co-operative where agents try to accomplish a goal as a team, or competitive where each individual agent tries to maximise their own benefit at the expense of others. Agents receive information from other agents and the environment and have internal rules that represent the cognitive decision process and determine how they respond. The rules can be a simple function of the inputs received or they can be very complex incorporating various internal state parameters, which can include a model representing the agent's world view of some part of the environment or even psychological models to include a kind of personality and emotional characteristics producing different agent behaviour under different circumstances (Trappl et al., 2003). Furthermore, these rules can either be fixed or they can change to represent learning. The system environment is susceptible to external pressure. This is a good model of a company where the environment becomes the organisation, the agents become managers and staff and external pressure might come from markets or customers.

So far we have discussed why agents and MAS are useful for us and we have learned what they are. In the next section we will take a look on how to build a software system based on intelligent agents.

MULTI-AGENT SYSTEM DESIGN

Designing simulations is not a trivial task, therefore well structured ways to design agents and multi-agent systems have been developed to guide the development. These comprise computer science design methodologies to develop the multi-agent system, the agent architecture, and some guidance about the format of the data required to model individual agent behaviour as well as for validating the multi-agent system.

Design Methodologies

In computer science a methodology is the set of guidelines for covering the whole lifecycle of system development both technically and managerially. It covers detailed processes for specifying, designing, implementing, testing/debugging as well as some guidelines for quality assurance, reuse of components, and project management. The ==design methodology== provides a process with detailed guidelines and notations that can be used to design the system, its components and the interactions between the components. The notation used for this is usually formal or semi-formal and can be of graphical or textual nature.

There are already object-oriented design methodologies (OODM) which one could argue would be useful for the task of designing MAS. But these methodologies have some flaws when used for MAS design. For example, as we mentioned earlier agents are pro-active, which means that they pursue their own goals. However, modelling of goals is not generally part of OODM. These methodologies have no means to model active objects. Therefore new methodologies have been developed to account for the specific needs of designing MAS. AOT Lab (2004) provides a good summary of the existing MAS design methodologies.

The following two sections will touch on the issue of representing reasoning which is something that is exceptionally well documented in the literature. Still it is sometimes difficult to transfer the theory into practice, especially if one is interested in a simple but effective representation of reasoning based on empirical data.

Agent Architectures

There are many ways to design the inner structure of an agent and many different agent architectures have been developed over the years. Wooldridge (2002) classifies architectures for intelligent agents into four different groups, as represented in Table 2. Furthermore, the table contains some examples of concrete agent architectures for each class with reference to some of their key contributors.

Table 2: Classification of agent architectures (after Wooldridge, 2002)

| Class | Examples of concrete architectures |
|---|---|
| Logic based agents | Situated automata (Kaelbing, 1986) |
| Reactive agents | Subsumption architecture (Brooks, 1986) |
| Belief-Desire-Intention agents | BDI architecture (Bratman et al 1988) |
| Layered architectures | Touring machines (Ferguson, 1992) |

Logic based agents are specified in a rule-like (declarative) language and decision making is realised through logical deduction. Reasoning is done off line, at compile time, rather than online at run time. The logic used to specify a logic based agent is essentially a modal logic of knowledge. The disadvantage of this architecture is that the theoretical limitations of the approach are not well understood. Through expressing the world as pure logic, it is arguable that concepts exist that have the potential to be unexplainable within such a rigid framework.

Reactive agents often use a subsumption architecture which is a way of decomposing complicated intelligent behaviour into many simple behaviour modules, which are in turn organised into layers. Each layer implements a particular goal of the agent, and higher layers are increasingly more abstract. Each layer's goal subsumes that of the underlying layers, for example the decision to move forward by the eat-food layer takes into account the decision of the lowest obstacle-avoidance layer. This architecture is often used when very robust agents are required capable of mimicking very complex adaptive behaviours

The Belief(s)-Desire(s)-Intention(s) architecture has its origin in practical reasoning, which means to decide what goals to achieve and how to achieve them. Beliefs represent the informational state of the agent or in other words its beliefs about the world, which may not necessarily be true and in fact may change in the future. Desires represent objectives or situations that the agent would like to accomplish or bring about, such as the possible goals. Intentions represent the deliberative state of the agent, such as the chosen goals. This architecture is often used when goals are not consistent, and therefore desires are a better way of expressing them. For example, the goals of saving money and buying items conflict with each other but might still be desirable.

The principal idea of layered architectures is to separate reactive, pro-active and social behaviour. These layers can be either horizontally organised where inputs and outputs are connected to each layer or vertical where each input/output is dealt with by a single layer.

Layered architectures are the most popular ones, although it is quite difficult to consider all the complex interactions between layers.

All of the described architectures are essentially blank canvases which require data to make them operational and enable the decision making process. The next section describes how different forms of behavioural information (theoretical or empirical) can be utilised for this purpose.

Internal Structure and Data Format

Developing intelligent agents requires gaining information about how agents make their decisions, how they forecast future developments, and how they remember the past. The challenge is to take real world information and make it useable for the agent to support the reasoning process in the most natural way. There are many design notations that allow you to describe the internal processes that run inside the agent during its reasoning process. In this section we will focus on describing the usage of state charts for this purpose.

State charts indicate what states an agent can be in and what triggers state changes for an agent. States are often defined through a delay time (the time it will take to finish a task) and the probability of occurrence. The central state of each state chart is the idle state which links to the all states that do not depend on a specific order of execution. Triggers can change the internal state or send a signal to other entities to change their state. States can have different levels of importance which impacts on when they are executed and if they are interruptible.

Knowledge incorporated within the state charts to trigger state changes can be represented in formulae, rules, heuristics, or procedures. Most often analysts only use one of these representation opportunities. The difficulty is that the knowledge has to be translated into a computational format, which requires extensive knowledge about how to do it. In order to define knowledge one could either use well established theoretical knowledge or collected empirical data. When using theoretical knowledge it is important to remember that it needs to fit in the application context in which it is to be used, a fact that is often ignored. Using empirical data requires a large number of high quality observations, either from the real system or derived through laboratory experiments (usually used to test very precise hypotheses). Sometimes there is a problem when parameters cannot be measured directly or estimated with sufficient precision. In this case, if historical data is available, one could also use the inverse method, such as using the output of a simulation to calibrate the inputs by comparing them to the output data of the real system (Brooks & Shi, 2006). The problem with this method is that there are usually many solutions and no method of identifying the correct one. Furthermore, there might be measurement errors contaminating the historical data.

Probabilities and distributions make a good choice as input data for companies and organisations. Often they can be built intuitively from existing business analysis data. Many companies collect data for making decisions based on statistics about the companies, markets, what's going on internally, and performance reviews. All of these are potential sources of information for building rules and distributions and to estimate the probabilities of events occurring. Another way to collect empirical data is through case study analysis. Based on the information from a specific system, with different types of information, ABS models can be developed. In such studies the analyst has multiple sources of qualitative and quantitative data, but incomplete information. Information from remote sensing, surveys, census data, field observation, and so on is often used to develop the different components of the system and to test the validity of the model. Often the goal is to understand the interactions between the different components of the system and to use the model to explore different policy scenarios.

In addition to these methods the agent approach allows for the use of some advanced techniques, such as neural networks, or genetic algorithms. Both are relevant to modelling dynamics of learning and adaptation, but out of the scope of this chapter.

AN EXAMPLE OF A MULTI-AGENT SYSTEM

In this section we provide a simple hypothetical example of a MAS to illustrate some of the key concepts discussed in the previous sections. The example encompasses the working of a small design department. Within this department there are two design teams each consisting of a supervisor and several designers. The department has one manager who interacts with the two supervisors. The designers interact amongst themselves and with the supervisor. The department receives contracts that define the activities required during a design project and respective deadlines. The design teams have to deliver the contracted design within the given time frame. The goal of the simulation study is to understand how team composition influences the productivity of a design department in fulfilling these contracts.

Relating this example back to the description of a MAS in the previous section:
- The design department is regarded as the environment in this model
- Contracts reflect inputs to the MAS as pressure from the outside as they have to be finished by certain deadlines
- People inside the department communicate with each other in a hierarchical way, the manager talks to supervisors and vice versa, supervisors talk amongst each other and in each design team designers talk to each other and their supervisor

Figure 1 shows the structure of the design department MAS. An optimal team composition is vital so that the contract productivity can be maximised. During a simulation run any agent member of the design department can evolve depending on who they work with. Consequently, their knowledge, communication and subsequently their productivity change throughout the simulation.

Figure 1: Structure of the design department MAS

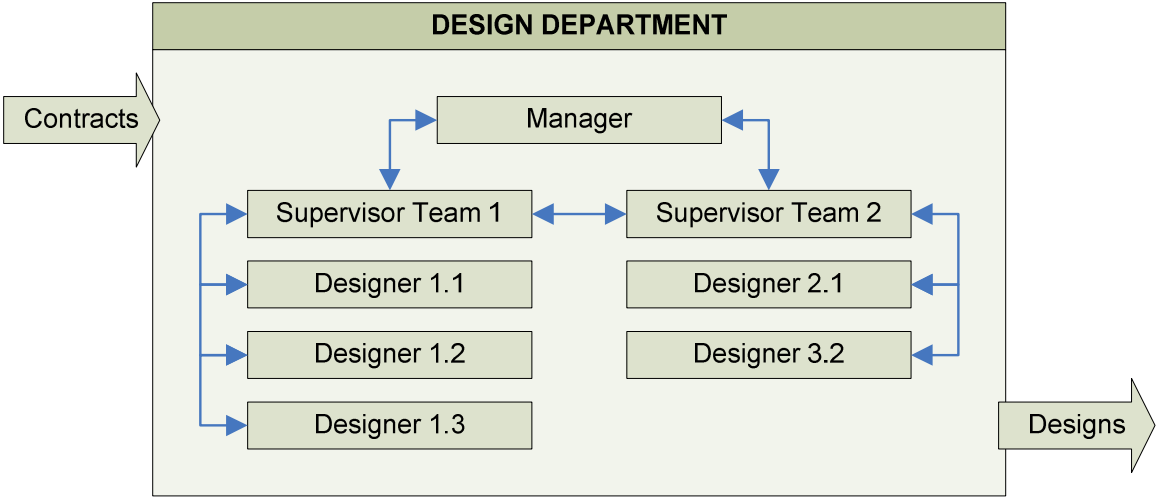

The MAS consists of heterogeneous agents. A reactive architecture for agents is used which responds to internal and external stimuli. As an example we will show how a designer agent could be modelled. Figure 2 shows the state chart for the designer agent. The states a designer agent can adopt are depicted by ovals and state changes are depicted by arrows. State changes can either be scheduled, for example project meetings, or they can be spontaneous, for example when receiving a request from another agent to provide support.

Figure 2: State chart for designer agents

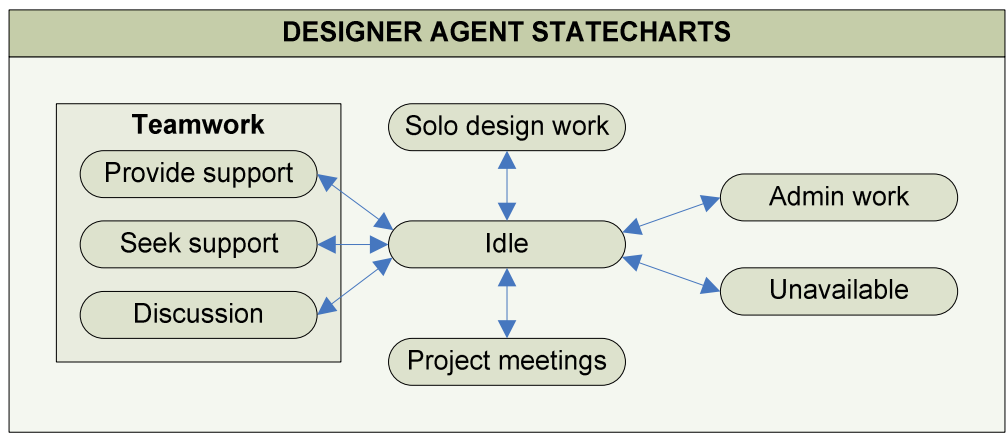

A designer agent has several attributes which evolve during the simulation and define the characteristics of the agents and consequently guide their behaviour and their decision making process:

- Knowledge (competency and technical skills of the designer)
- Communication (how good the designer can expresses their knowledge)
- Productivity (how efficient the designer is in what they are doing)

Knowledge can be divided into sub-categories, for example planning, design, and testing. For different tasks or different stages of the project different knowledge requirements will be set. If the designer agent has been allocated a task but does not possess the required knowledge, they seek support from a colleague with the correct knowledge capability and hope that a colleague with the right capabilities is available and willing to provide support. In addition to the attributes each designer agent has several fixed parameters which define the agent's stereotype at the beginning of the simulation execution:

- Start-up values for the attributes

- Personality traits (e.g. willingness to support; willingness to communicate)

Attribute changes of an agent depend on the current attribute values, the activities they are performing and the colleagues they are working with. For example: Project A consists of one supervisor and three designers with high knowledge but poor communication skills. If the project requires a lot of teamwork, productivity of the team will be very low. Adding a good communicator to the team will improve the communication skills of the team and increase productivity but the communication skills of the added team member will decline, due to potential motivational issues such as frustration.

To make the simulation more realistic some form of cost could be added, for example people with extraordinary skills cost the company more than those with average skills.

RESOURCES

This section will provide you with resources which should be useful if you want to get actively involved in ABS. For further reading about the subject we would recommend Weiss (1999). The book provides detailed coverage of basic topics as well as several closely related ones and is written by many leading experts, guaranteeing a broad and diverse base of knowledge and expertise. It deals with theory as well as with applications and provides many illustrative examples.

Software Resources

Due to the growing interest in agent-based modelling and simulation and the fact that it is particularly popular in academia there are now many sophisticated free toolsets available.

Some of them are equipped with a visual interactive user interface and therefore relatively easy to use, even for non-professional programmers. An example of such development environments are Repast Symphony (Repast, 2007). There are also some professional development environments available, for example AnyLogic (XJ Technologies, 2007). A well maintained list of software resources can be found at SwarmWiki (2007).

Formal Bodies, Discussion Groups, and Organisations

The Foundation of Intelligent Physical Agents (FIPA) is a world-wide, non-profit association of companies and organisations that promotes agent-based technology and the interoperability of its standards with other technologies (FIPA, 20007). Although FIPA is intended to deal with physical agents (e.g. for robotics or agent web services) many standards are also used to design non-physical MAS.

A very active mailing list devoted entirely to MAS is the mailing list by Jose Vidal (Vidal, 2007). It provides discussions and announcements about MAS. There is also a website linked to this discussion group with a wealth of information about MAS. SimSoc (JISCmail, 2007) is a discussion group linked to the Journal of Artificial Societies and Social Simulation (JASSS, 2007) which is freely available on the internet. The discussion group offers news and discussions about computer simulation in the Social Sciences.

The website of the Society for Modelling and Simulation International (SCS, 2007) is a good place to look for conferences and workshops, as is the INFORMS Simulation Society website (INFORMS, 2007). In addition the INFORMS offers the proceedings of the Winter Simulation Conferences they organise on their website free of charge. This is a prime information resource for everyone who is interested in the latest developments in the world of simulation.

CONCLUSION

In this chapter we have discussed simulation as a useful tool for the analysis of organisations. If the individual elements within the organisation to be modelled are of a heterogeneous and pro-active nature, for example staff members or customers, ABS seems like a well suited analysis tool. Formal methods exist to develop and validate such simulation models. This ensures confidence in the results and underpins the value of ABS as a support tool in the strategic and tactical decision making process.

With this overview and insight into the usefulness of ABS models, the reader is well placed to start developing and modelling their own ideas.

TERMS AND DEFINITIONS

**Agent-based simulation**: A bottom up approach for modelling system changes over time. In an agent-based simulation model the researcher explicitly describes the decision process of simulated actors at the micro level. Structures emerge at the macro level as a result of the actions of the agents and their interactions with other agents and the environment.

**Artificial white-room**: Simulation of a laboratory as it is used by Social Scientists for data gathering under controlled conditions.

**Discrete event simulation**: Modelling a system as a set of entities being processed and evolving over time according to availability of resources and the triggering of events. The simulator maintains a queue of events sorted by the simulated time they should occur.

**Emergent behaviour**: Refers to the way complex systems and patterns of behaviour develop out of a multiplicity of relatively simple interactions amongst agents and between agents and their environment over a certain period of time.

**System dynamics**: A top down approach for modelling system changes over time. Key state variables that define the behaviour of the system have to be identified and these are then related to each other through coupled, differential equations.